\documentclass[10pt,twocolumn,letterpaper]{article}

\usepackage{cvpr}
\usepackage{times}
\usepackage{epsfig}
\usepackage{graphicx}
\usepackage{subcaption}
\usepackage{caption}
\usepackage{amsmath}
\usepackage{amssymb}
\usepackage{eucal}

\usepackage[breaklinks=true,bookmarks=false]{hyperref}

\cvprfinalcopy 

\def\cvprPaperID{****} 

\begin{document}

\title{ TunaGAN: Interpretable GAN for Smart Editing}

\author{Weiquan Mao\footnotemark[1] \qquad Beicheng Lou\footnotemark[1] \qquad Jiyao Yuan\footnotemark[1]  \\
Stanford University\\
{\tt\small mwq@stanford.edu} \qquad {\tt\small lbc45123@stanford.edu} \qquad {\tt\small yuan999@stanford.edu} 
}

\maketitle

\renewcommand{\thefootnote}{\fnsymbol{footnote}}
\footnotetext[1]{These authors contributed equally to this study and share first authorship.}

\begin{abstract}
   In this paper, we introduce a tunable generative adversary network (TunaGAN) that uses an auxiliary network on top of existing generator networks (Style-GAN) to modify high-resolution face images according to user's high-level instructions, with good qualitative and quantitative performance. To optimize for feature disentanglement, we also investigate two different latent space that could be traversed for modification. The problem of mode collapse is characterized in detail for model robustness. This work could be easily extended  to content-aware image editor based on other GANs and provide insight on mode collapse problems in more general settings.
\end{abstract}

\section{Introduction}
Generative adversary network (GAN) has proved successful in learning the latent representation of faces and generating realistic-looking images \cite{nipsgan}. There has also been substantial effort in using the generative model for content-aware editing \cite{infogan,interactivegan,introspectivegan, discogan}. Ideally, one hopes to capture semantically meaningful features (e.g. smile, gender, beard etc.) in a latent vector so that these features in the output image can be tuned by changing the latent vector linearly. Previous work typically involves inputting such a latent vector along with noise to generator, where feature disentanglement in the latent vector space is encouraged during training. Style-GAN \cite{stylegan} is the latest state-of-art artificial face generator with unprecedented resolution and decent feature disentanglement, which makes a good candidate for exploitation for content-aware editing of facial images.

\begin{figure}[h]
\begin{center}
   \includegraphics[width=0.7\linewidth]{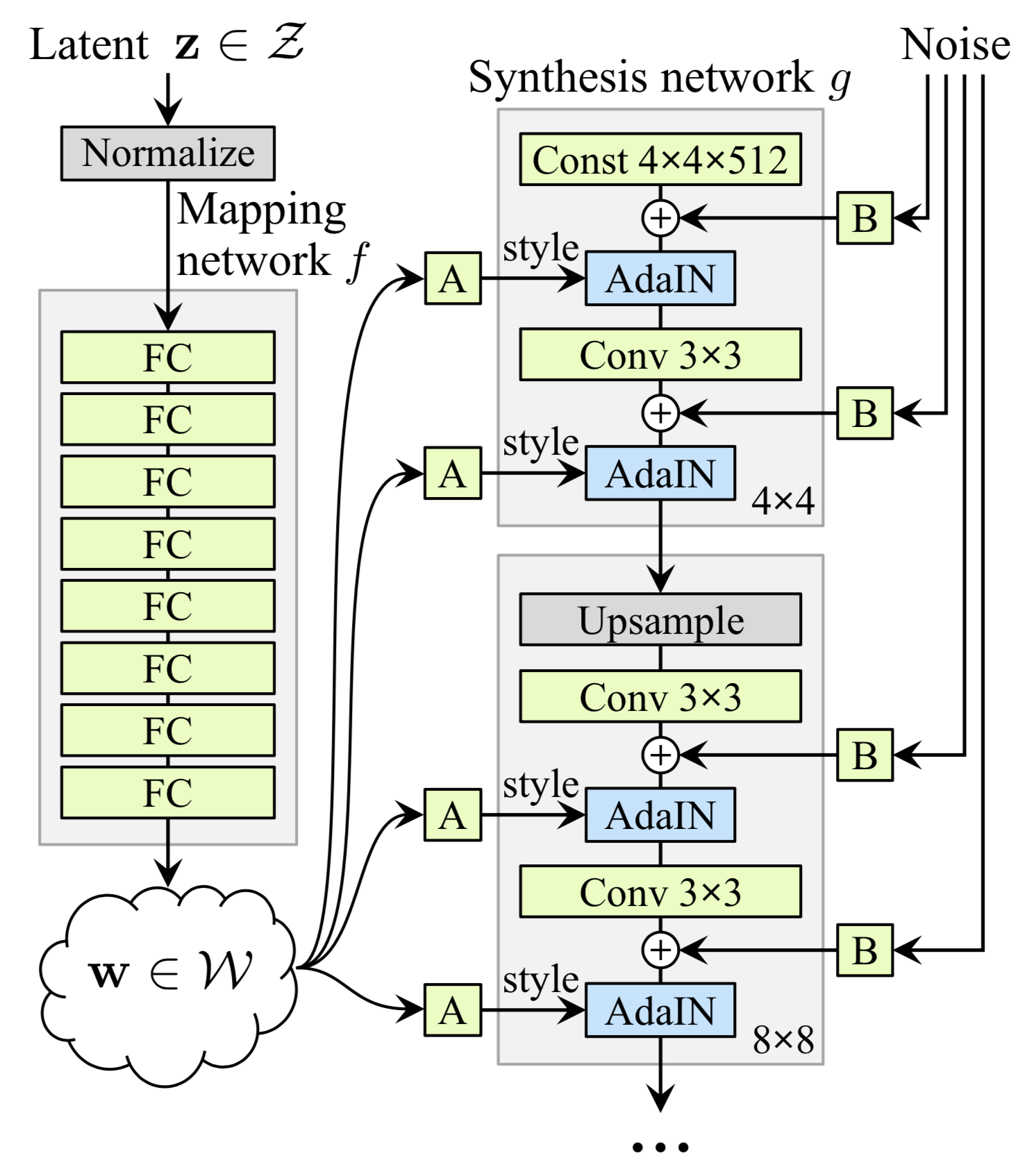}
\end{center}
   \caption{Architecture  of Style-GAN}
\label{fig:stylegan}
\end{figure}

However, the training process takes weeks even on NVDIA's powerful hardware, which makes adapting their network formidable for researchers. As a result, for us to exploit such a trained network, instead of modifying the architecture, it is more efficient to use the trained network as a modular block together with other small customized networks to achieve the desired input-output mapping. 

This project investigates the possibility and performance of a content-aware editing network based on Style-GAN.

We aim to implement a robust image editing tool with high flexibility which takes an image (of face) and the wanted high-level modification instructions (e.g. more masculine, less beard) as input, and outputs a natural-looking modified image.

\begin{figure}[h]
\begin{center}
   \includegraphics[width=0.8\linewidth]{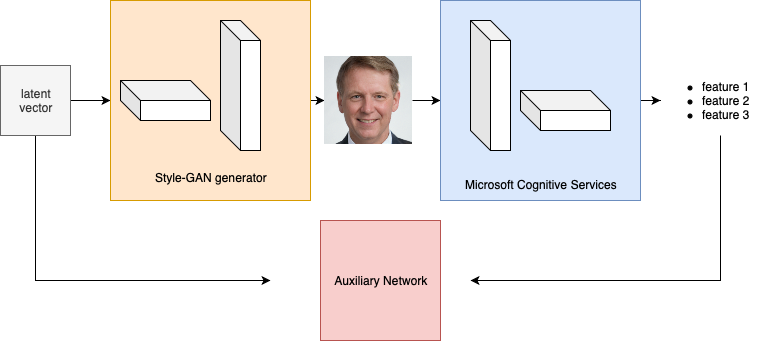}
\end{center}
   \caption{Architecture of TunaGAN}
\label{fig:tunagan}
\end{figure}

In formal notation consistent with \cite{stylegan} (Figure \ref{fig:stylegan}), our implementation takes an input image $x$ (with features encoded as $y$) and the wanted modification $\Delta y$ (which can be selected from a pool or obtained by image comparison), and then outputs the modified image $\hat{x}$ with features encoded as $y+\Delta y$. By utilizing Style-GAN to map a latent vector $z$ (or $w$) to image $x$, we reduce the problem to finding the correct $\Delta z$ (or $\Delta w$) that results in $\Delta y$, as captured by the pink box in Figure \ref{fig:tunagan}. We coined the name 'TunaGAN' for this process of making an existing GAN tunable by blending it with other networks.

Along the implementation, we encountered mode collapse when traversing the latent vector space near zero. We characterize it with visualizations and provide detailed explanations.


\section{Related work}

During past few years, interesting progress has been made in using GANs to generate high-resolution artificial images. Tero Karras\cite{pggan} put forward a training methodology for GANs where they started with low-resolution images, and then progressively increased the resolution by adding layers to the networks. This method (PG-GAN) speeds the training up and greatly stabilizes it, allowing us to produce images of high quality. Style-GAN \cite{stylegan} proposed an alternative generator architecture for generative adversarial networks, which leads to an automatically learned, unsupervised separation of high-level attributes. 

Based on PG-GAN or Style-GAN's fake-face generator, in order to apply our model to real-life images, we need to inversely transform real-face images to latent vector representations so that the generated images are like the original reference face. Several works attempt at recovering the latent representation of an image with respect to a generator. From a theoretical perspective, Bruna \cite{bruna2013signal} explore the theoretical conditions for a network to be invertible. We are going to introduce a neural network to map real-face images to latent representations below. The core idea is inspired by Justin Johnson \cite{johnson2016perceptual}. They put forward to use the perceptual loss functions for training feed-forward networks for image transformation tasks. Especially for super-resolution, their method trained with a perceptual loss is able to better reconstruct fine details compared to methods trained with per-pixel loss.

There has been rapid advancement in modeling the image manifold. Jun-Yan Zhu \cite{naturalimagemanifold} used GAN as a constraint on the output of various image manipulation operations, to make sure that the results lie on the learned manifold at all times. This idea enables them to reformulate several editing operations, specifically color and shape manipulations, in a natural and data-driven way. The model automatically adjusts the output keeping all edits as realistic as possible and shows that it is possible to recover z from a generated sample. Radford and Metz added a set of constraints to improve the stability of training GANs to create deep convolutional generative adversarial networks (DC-GAN)\cite{DC-GAN}. DC-GAN also saw interesting results with arithmetic properties on faces. For example, it showed that $smiling~woman - neutral~woman + neutral~man$ produced the image of a smiling man. 

The latent space of GANs seems to linearize the space of images. That is: interpolations between a pair of z vectors in the latent space map through the generator to a semantically meaningful, smooth nonlinear interpolation in image space. Piotr\cite{bojanowski2017optimizing} introduced Generative Latent Optimization (GLO), a framework to train deep convolutional generators using simple reconstruction losses. The model interpolates between examples that are geometrically quite different, reconstructing the rotation of the head from left to right, as well as interpolating between genders or different ages. This idea has been used in our project as well. Bo\cite{zhao2018modular} constructed ModularGAN for multi-domain image generation. It has superior flexibility of generating an image in any desired domain.

It would be good if we know GANs at the unit-level or object-level so that we can manipulate GANs to edit image with high-level modification instruction. The work of David and Jun-yan\cite{bau2018gan} inspires us. They presented a general method for visualizing and understanding GANs at different levels of abstraction, from each neuron, to each object, to the contextual relationship between different objects. With their method, we can interactively manipulate objects in a scene.

\section{Dataset}
\begin{figure}[h]
\begin{center}
   \includegraphics[width=\linewidth]{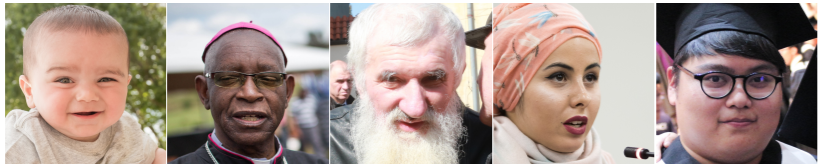}
\end{center}
   \caption{Some samples from FFHQ dataset}
\label{fig:data_sample}
\end{figure}

For pre-trained Style-GAN, we use Nvidia's new dataset of human faces (Flickr-Faces-HQ, FFHQ) that offers much higher quality and covers wider variation than existing high-resolution datasets. The dataset consists of 70,000 PNG images at $1024^2$ resolution and contains considerable variation in terms of age, ethnicity and image background. Some samples are shown in figure \ref{fig:data_sample}. The data is split into $85.7\%$ and $14.3\%$ for training and validation respectively.

For auxiliary neural network, we firstly use 20307 random $z$ latent vectors to generate 20307 $w$ latent vectors using an 8-layer multilayer perceptron (MLP) in pretrained Style-GAN, where both $z$ and $w$ latent vectors serve as the input of auxiliary neural network. The dimensions of $z$ and $w$ are 512 and 9216 respectively. We then input 20307 $z$ latent vectors and use the Style-GAN to generate 20307 fake human faces images at $256^2$ resolution. Those images are then labelled by Microsoft Cognitive Services API, where 27 face attribute features are available such as age, emotion, gender, pose, smile, and facial hair. With the data, we split it into $80\%$ for training and $20\%$ for validation.

\section{Methods} \label{methods}

Our baseline model is Style-GAN (Figure \ref{fig:stylegan}), which has an innovative style-based generator where a latent vector $z \in \mathcal{Z}$ is firstly mapped by a non-linear network $f$ to an intermediate latent space $\mathcal{W}$ before being fed to synthesis network $g$. The dimension of $z$ is set to $512$ and $w$ to $18\times512$. Learned affine transformation maps $w$ to styles $y = (y_s, y_b)$ that control adaptive instance normalization (AdaIN) operations after each convolution layer of the synthesis network. The generator is also provided with explicit noise inputs for stochastic detail. The discriminator or the loss function is not modified compared to previous work.

In spite of the state-of-art performance, interpretability remains a problem for Style-GAN, which makes it hard to tune specific features of the generated image. However, the projection of feature change onto latent space is typically continuous, meaning that the interpolation between two vectors in the latent space usually leads to a smooth transition of corresponding images in the feature space. Therefore, by correctly traversing the latent space, we can still achieve feature tuning based on Style-GAN, circumventing the need for interpretability.

\subsection{Map image to latent vector} \label{methods-map}

In order to manipulate real-face images instead of fake-face images generated by GANs, we first try to map image to latent vector so that we can generate images similar to the reference one. We train a neural network with discrepancy in feature space as loss. Namely, we first extract features of the referenced image $\hat{x}$ from VGG-16 at the layer 9 as $\hat{y}$. And the generated image with latent vector $\hat{w}$ can produce a feature vector $y$ in the same way. Then we minimize $(\hat{y}-y)^2$ to find the latent vector $\hat{w}$ corresponding to image $\hat{x}$. 

\begin{figure}[h]
\begin{center}
   \includegraphics[width=0.5\linewidth]{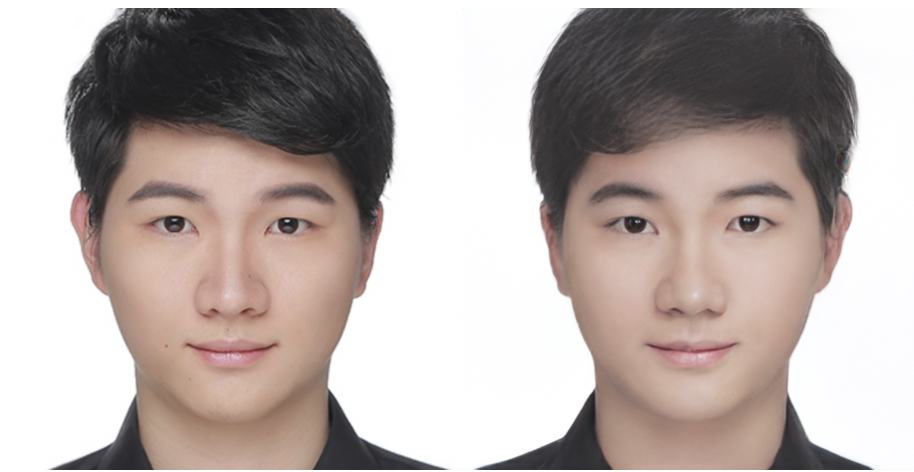}
\end{center}
   \caption{Reconstruction of image from optimal latent vector $\hat{w}$}
\label{fig:reconstruction}
\end{figure}

This mapping is in principle not ideal, but we can reproduce the image $\hat{x}$ from $\hat{w}$ to a satisfying extent, as shown in Figure \ref{fig:reconstruction}. The left image is real and the right image is reconstructed from latent vector, which has little difference with reference one.

\subsection{Traversing in latent space} \label{methods-traverse}

To modify latent vector correctly for wanted feature change, we first fit the mapping from latent space to feature space by support vector machine (SVM) or neural network (NN). Then optimal trajectory in the latent space for tuning certain features could be calculated by running optimization on latent vector as trainable variable. To obtain training data, we input latent vector to Style-GAN to produce synthetic image which is sent through Microsoft Cognitive Services API for feature extraction. With the mapping from latent space to feature space, we could compute the latent vector direction in either $\mathcal{Z}$ or $\mathcal{W}$ space corresponding to wanted feature change.
 
Several architectures have been studied in our experiment. To traverse latent space linearly, logistic regression is used to classify categorical features like gender and glasses while linear regression is used to predict numerical features like age and smile. We notice that the features could not be accurately separated by linear model, thus experiment with a two-layer NN as our feature extractor with dropout as regularization. A trick here is that we implement another embedding network so that finding the correct trajectory in latent space can be converted to a gradient descent problem.
 
Finally, with the obtained latent vector directions $\Delta z$ or $\Delta w$ corresponding to wanted feature changes, we can input $z + \Delta z$ or $w + \Delta w$ to Style-GAN to generate new images.


\section{Results}

\subsection{Preliminary results} \label{preliminary}
\begin{figure}[h]
\begin{center}
   \includegraphics[width=\linewidth]{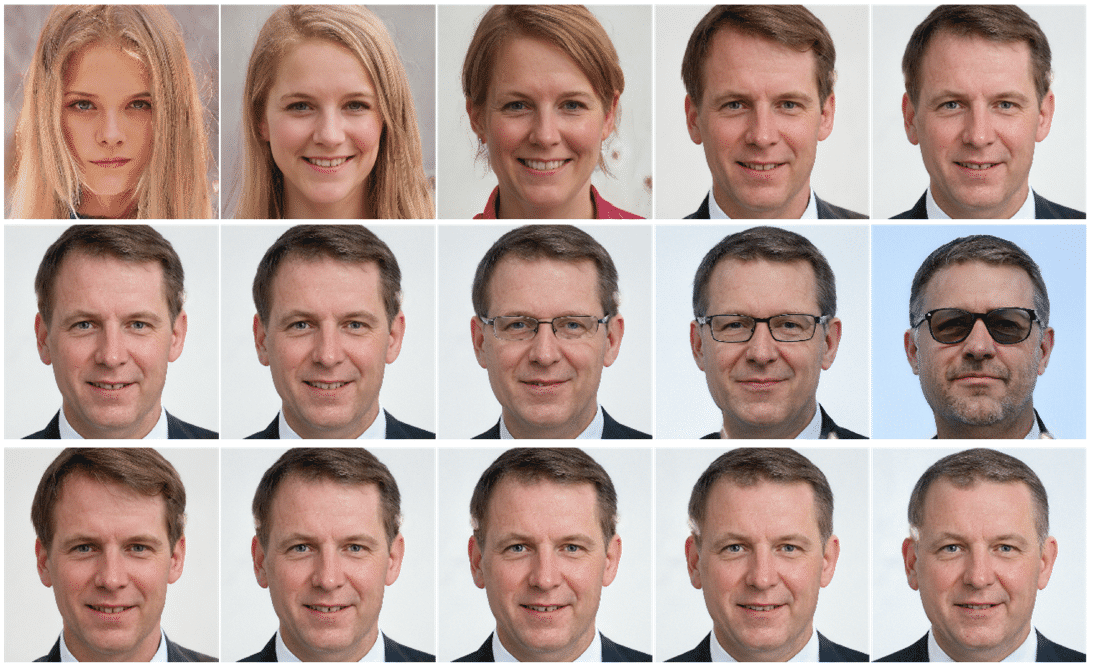}
\end{center}
   \caption{Demonstration of qualitative results by tuning $z$}
\label{fig:demo}
\end{figure}

To illustrate the effectiveness of latent vector tuning, we map the wanted feature modification $\Delta y$ to optimal latent vector change $\Delta z$. By traversing latent space $\mathcal{Z}$ linearly, we obtain satisfying qualitative results of variation along 'gender', 'beard', 'hair' features as shown in Figure \ref{fig:demo}.

\begin{figure}[h]
\begin{center}
   \includegraphics[width=\linewidth]{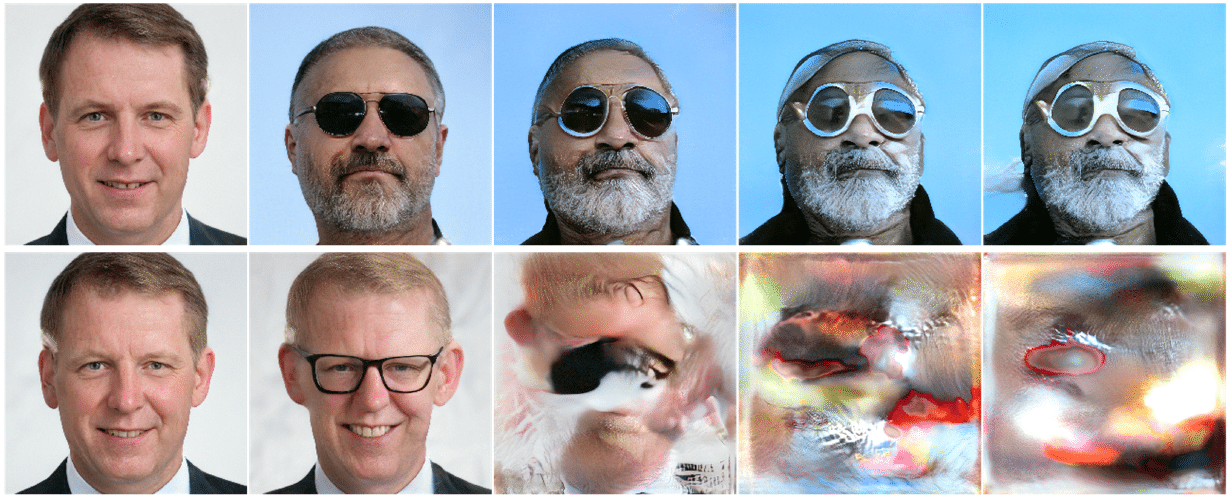}
\end{center}
   \caption{Traversing $\mathcal{Z}$ (upper) vs $\mathcal{W}$ (lower) latent space}
\label{fig:wvsz-linear}
\end{figure}

We further investigate the differences between traversing $\mathcal{Z}$ and $\mathcal{W}$ latent space by firstly exploring the model stability when tuning latent vector linearly with large step. As shown in the second row of Figure \ref{fig:wvsz-linear}, generated images achieved by traversing $\mathcal{W}$ space is no more like faces once step size gets large. In contrast, equally large step size in $\mathcal{Z}$ space results in well-behaved output even in the saturation regime. The reason is that the 8-layer MLP in Style-GAN ensures that $z$ with large $\Delta z$ maps to $w$ with relatively small $\Delta w$. However, the same scale of changes in $\mathcal{W}$ space is more likely to cause non-face images being generated. 

\begin{figure}[h]
\begin{center}
   \includegraphics[width=0.7\linewidth]{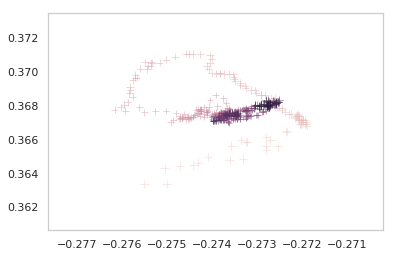}
\end{center}
   \caption{Ideal traversing according to nonlinear model from no beard (light) to heavy beard (dark)}
\label{fig:gradient}
\end{figure}

Furthermore, we notice that feature disentanglement is still not perfect in Style-GAN and speculate that we can improve the network by traversing in latent space in a nonlinear way. Figure \ref{fig:gradient} illustrates how the ideal traversing in latent space is highly nonlinear. Here we visualize the trajectory in latent space when we smoothly strengthen the feature 'beard'. The plotted axes correspond to the direction with largest gradient in latent space.

\begin{figure}[h]
\begin{center}
   \includegraphics[width=\linewidth]{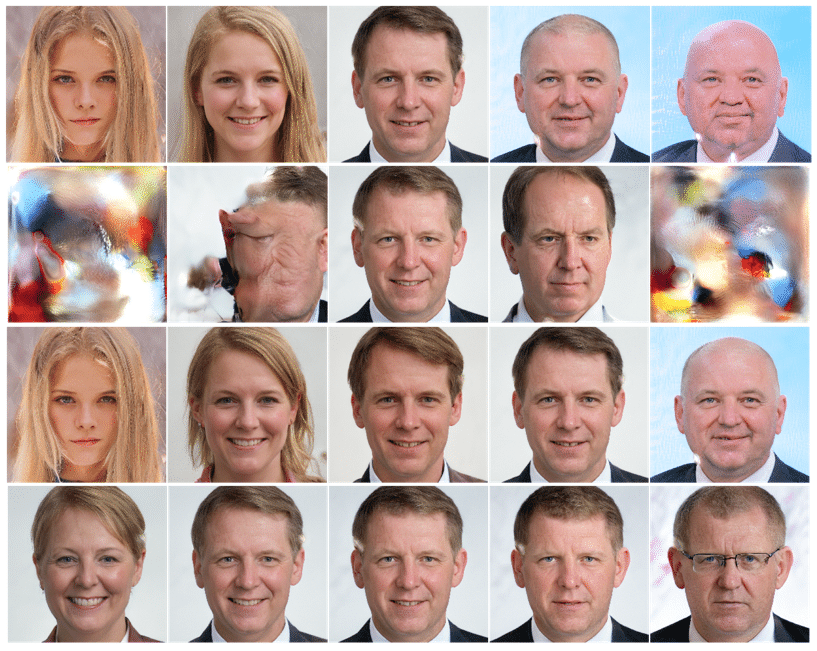}
\end{center}
   \caption{Comparison among different models (up to down: linear z, linear w, nonlinear z and nonlinear w)}
\label{fig:four_methods}
\end{figure}

Thus we try traversing both $\mathcal{Z}$ and $\mathcal{W}$ spaces non-linearly. We firstly train a 2-layer neural network, denoted as $f_n$, to map latent vectors to features. $f_n$ could be designed deeper for better performance. An embedding layer $f_e$ is then concatenated before $f_n$ to help obtain the object latent vector. As show in figure \ref{fig:four_methods}, the performance of nonlinear models for both $z$ (third row) and $w$ (fourth row) are better than linear models for $z$ (first row) and $w$ (second row) respectively. The stability of traversing in $\mathcal{W}$ space has been significantly improved.

\begin{figure}[h]
\begin{center}
   \includegraphics[width=\linewidth]{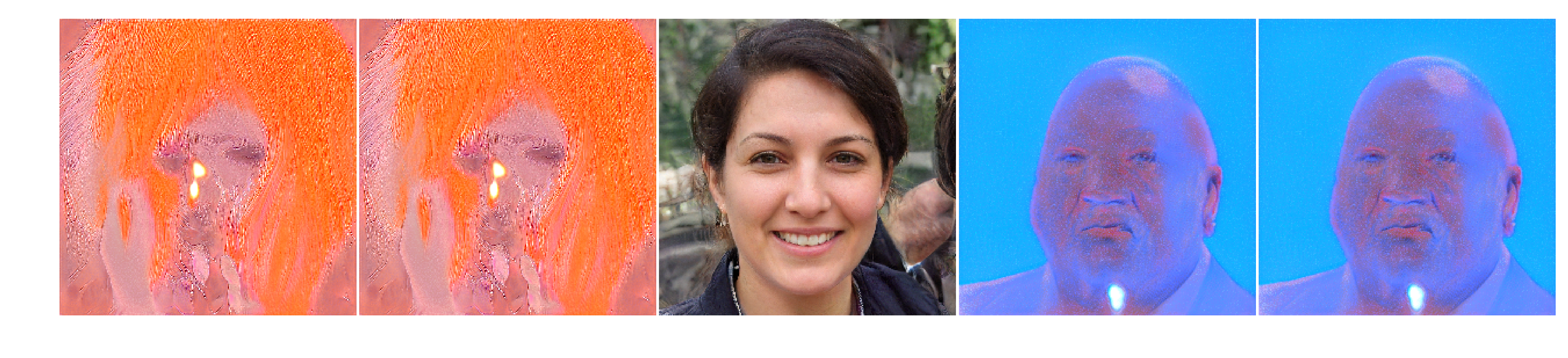}
\end{center}
   \caption{Mode collapse for traversing $\mathcal{Z}$ latent space}
\label{fig:zexplosion}
\end{figure}

While traversing $\mathcal{Z}$ latent space is more stable than in $\mathcal{w}$ space when large steps are taken, there are still certain inputs that will cause collapse for traversing in $\mathcal{Z}$, as shown in Figure \ref{fig:zexplosion}. Here we deliberately choose the initial image that corresponds to $z=0$. The results indicate that even a small change of latent vector direction will crash the generated image. More observations and explanations of mode collapse will be introduced in section 5.5.

To avoid the mode collapse when traversing $\mathcal{Z}$ space, traversing $\mathcal{W}$ is preferred. Due to the stronger disentanglement of $\mathcal{W}$ space (compared to $\mathcal{Z}$ space), it is expected that the $f_n$ of nonlinear $\mathcal{W}$ model has a simpler structure which indicates the feasibility to train a good-performance $f_n$. In our experiment, even a shallow 2-layer fully connected neural network is capable to ensure the accuracy of more than $90\%$ for categorical features such as beard, gender and glasses. In comparison, the same structure of $f_n$ for nonlinear $\mathcal{Z}$ model could only reach accuracy at around $76\%$.

However, high accuracy of $f_n$ does not necessarily ensure the overall performance of nonlinear model. For example, when data is highly unbalanced, predicting the majority class results in high accuracy of $f_n$ but poor performance of vector tuning since no information of how features varies in latent space is provided by $f_n$. In our experiment, the accuracy of $f_n$ for predicting beard reaches $95\%$ in both $\mathcal{W}$ and $\mathcal{Z}$ spaces but the quality of generated images by both models is worse than other attributes tuning. Furthermore, unbalanced data will cause generated images crashing easily with larger tuning coefficients using nonlinear $\mathcal{W}$ model.

We then investigate how number of layers of $f_n$ affect the performance of our model. We concatenate different numbers of fully connected layers with a two-layer shallow network and evaluate the performance by observing both the accuracy of $f_n$ and images quality when tuning latent vectors using auxiliary network. Theoretically speaking, when adding a 8-layer MLP (same as in Style-GAN) in nonlinear $\mathcal{Z}$ model and keeping parameters of MLP fixed, it is equivalent to a nonlinear $\mathcal{W}$ model. Thus it is theoretically feasible to further improve the nonlinear $\mathcal{Z}$ model by tuning parameters of MLP which could outperform than current nonlinear $\mathcal{W}$ model. However, practically speaking, training models in $\mathcal{W}$ space saves the troubles in tuning parameters (compared to $\mathcal{Z}$) and ensures the satisfying performance.

\subsection{Quantitative metric}
To quantitatively evaluate the performance of our model, we use Fr$\acute{e}$chet inception distances (FID), inception score (IS) and separability score (SS). Both FID and IS reflect the quality of generated images, which are the same as \cite{stylegan}.

To estimate disentanglement of latent space, we use SS, which measures the separability of vector space. SS has minimum value 0, which means perfect separability; whereas, higher value reflects less separability. We compute the conditional entropy between true labels and predicted labels by linear and nonlinear models. We use 3000 and 20000 (latent vector, feature) pairs to train SVM (linear) and shallow neural network (nonlinear) respectively. We calculate the final score as exp$(H(Y|X))$, where we take gender, beard and glasses as sample attributes.

The result is shown in Table \ref{table:SS}. It is as expected that $\mathcal{W}$ space has lower SS than $\mathcal{Z}$ since 8-layer MLP in Style-GAN is proved to have successfully mapped $\mathcal{Z}$ space into a more disentangled $\mathcal{W}$ space. The higher disentanglement of $\mathcal{W}$ space could also be reflected from the architecture of auxiliary neural network, where shallow networks for $w$ could gain better performance than even deeper networks for $z$. It is also shown from Table \ref{table:SS} that nonlinear model has lower SS than linear model since direction for nonlinear model is not only dependent on attributes but the latent vector itself.

\begin{table}[!htb]
\begin{tabular}{|c|c|c|c|c|}
\hline
\textbf{Model}       & \textbf{Gender} & \textbf{Beard}  & \textbf{Glasses} & \textbf{Overall}    \\ \hline \hline
Linear $\mathcal{Z}$    & 2.2057 & 1.3243 & 1.8572  & 5.4249 \\ \hline
Linear $\mathcal{W}$    & 1.8984 & 1.2921 & 1.5709  & 3.8533 \\ \hline
Nonlinear $\mathcal{Z}$ & 1.6361 & 1.0979 & 1.3946  & 2.5051 \\ \hline
Nonlinear $\mathcal{W}$ & 1.3079 & 1.1335 & 1.1384  & 1.6877 \\ \hline
\end{tabular}
\\
\caption{Separability score for different models}
\label{table:SS}
\end{table}

\subsection{Applied to real image}
\begin{figure}[h]
\begin{center}
   \includegraphics[width=0.9\linewidth]{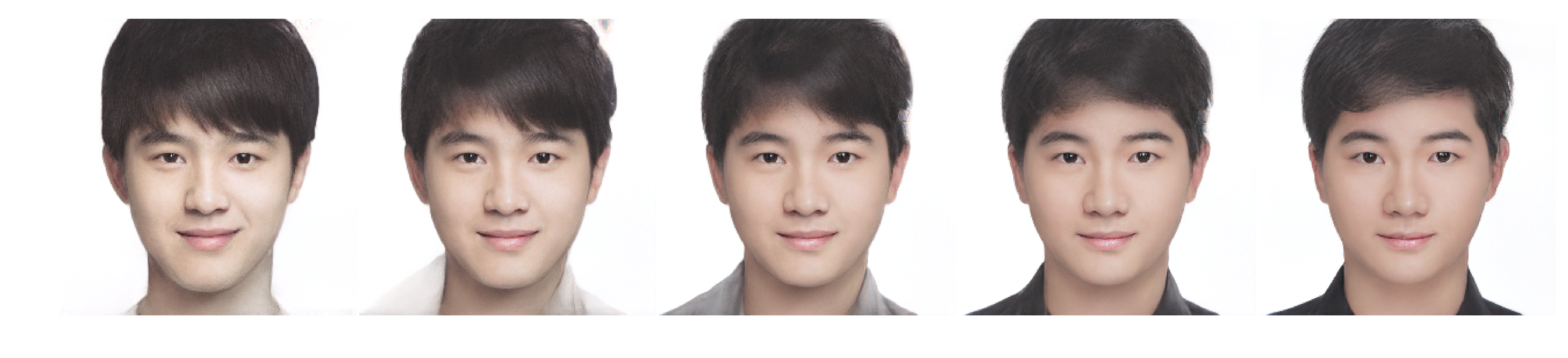}
\end{center}
   \caption{Demonstration of interpolation of two real-life images, where interpolation trajectory can be specified arbitrarily in feature space}
\label{fig:demo-interpolation}
\end{figure}

Section \ref{preliminary} demonstrates the effectiveness of tuning latent vector for wanted feature change. The images shown there, albeit realistic-looking, are all generated from arbitrary latent vector and tuned away thereafter. For real-life applications, we will need to map real image to latent vector, modify it and then map it back to image, as described in Section \ref{methods}.
Figure \ref{fig:demo-interpolation} demonstrates interpolation between two images, as is also shown in \cite{stylegan}. If wanted, our method allows more flexible interpolation trajectories by first mapping both images to feature space and then map the feature-space interpolation back to real images.

Figure \ref{fig:demo-interpolation} demonstrates modification of real-life image by mapping it to latent vector, traversing latent space correctly and mapping it back to image.
\begin{figure}[h]
\begin{center}
   \includegraphics[width=0.9\linewidth]{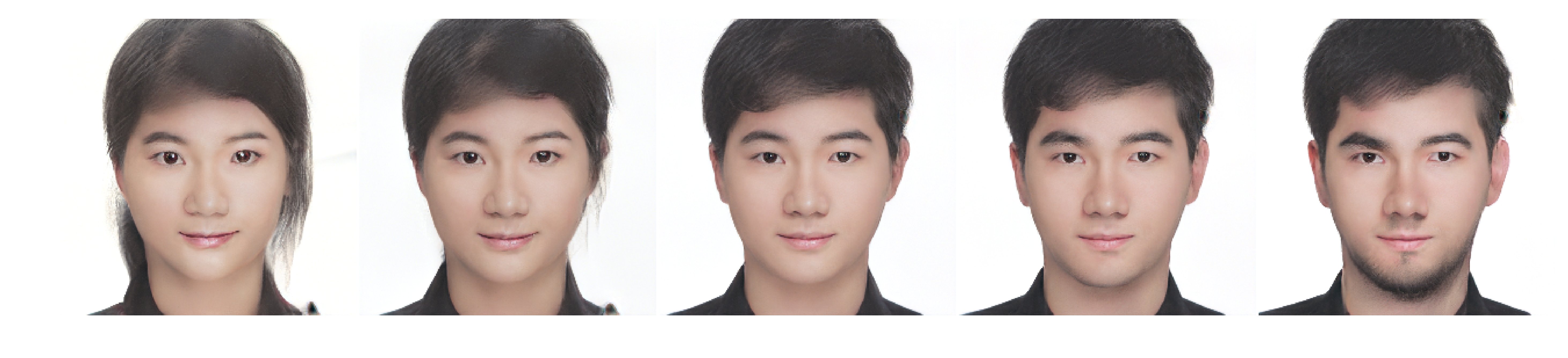}
\end{center}
   \caption{Demonstration of tuning of real-life image (modifying gender)}
\label{fig:demo-tuning}
\end{figure}

\subsection{Study of mode collapse}
As shown in Figure \ref{fig:zexplosion}, if the starting image corresponds to a latent vector of all zeros, modification in $\mathcal{Z}$ latent space will lead to mode collapse even for very tiny step size. Namely, trying to make the figure slightly more masculine will immediately result in the monster-like image of the bald fat man that captures the machine's stereotype of the concept 'male'. Understanding this problem is both practically important (so that this lucky woman can modify her image with our method) and theoretically interesting (to shed light on mode collapse in a concrete setting).

\begin{figure}[h]
\begin{center}
   \includegraphics[width=0.9\linewidth]{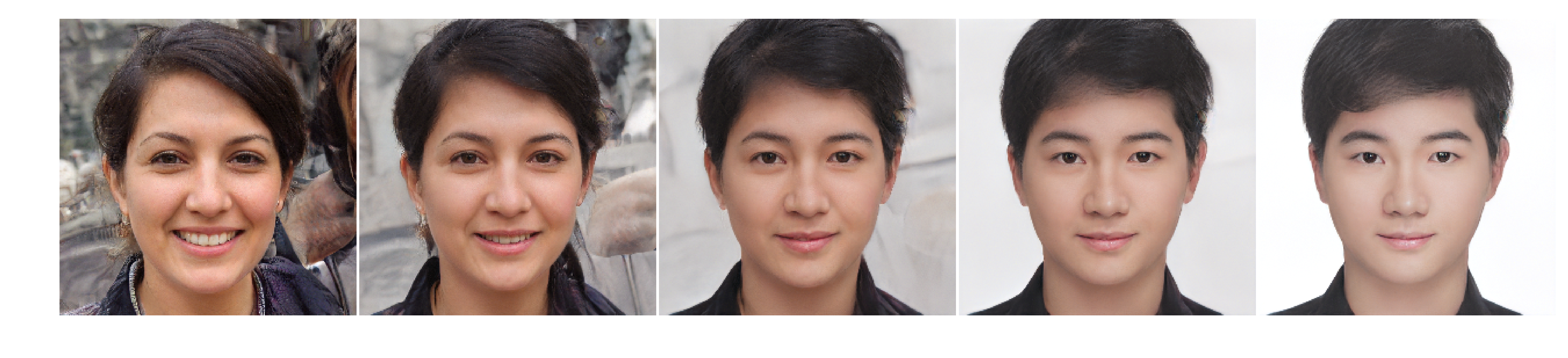}
\end{center}
   \caption{Traversing near zero in $\mathcal{Z}$ space towards other direction}
\label{fig:mode-mwq}
\end{figure}

To understand this mode collapse, we first note that traversing in $\mathcal{Z}$ latent space from zero towards more male is effectively interpolation between the image corresponding to zero latent vector (middle of Figure \ref{fig:zexplosion}) and the image of machine's stereotype of 'male' (rightermost of Figure \ref{fig:zexplosion}). This interpolation saturates immediately when we move away from zero. However, if we move from zero towards some other direction, the interpolation does not immediately saturate. In Figure \ref{fig:mode-mwq}, the image smoothly transforms between the two ends.

\begin{figure}[h]
\begin{center}
   \includegraphics[width=0.95\linewidth]{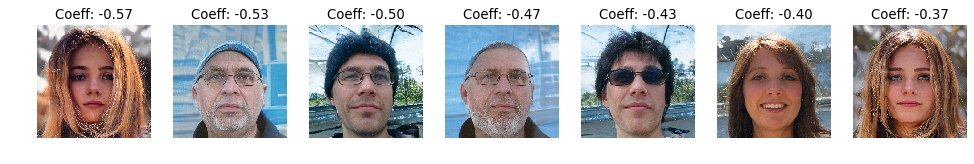}
\end{center}
   \caption{Oscillation near zero in $\mathcal{Z}$ space for small step size (left to right: increasingly male)}
\label{fig:mode-oscillation}
\end{figure}

This confirms the intuition described in Section \ref{preliminary} that the direction in latent space for feature change is an unstable mode of the network. A simple-minded solution would be to decrease the step size further. However, as shown in Figure \ref{fig:mode-oscillation}, by traversing latent space towards more 'female', the image does not monotonically transforms towards more 'female'. Instead, the image oscillates between 'male' and 'female'. Furthermore, the traversing clearly entangles other features like 'glasses'.

This can be understood from the non-normalizability of zero vector. All $\mathcal{Z}$ vectors are normalized to norm 1 before being passed through the fully connected layers, except for zero vector which will remain zero. Since it is inevitable for the latent space traversing to be noisy (i.e. the direction is not perfectly correct), the starting vector having a norm of zero would make all such noise appear relatively huge. 

\begin{figure}[h]
\begin{center}
   \includegraphics[width=0.7\linewidth]{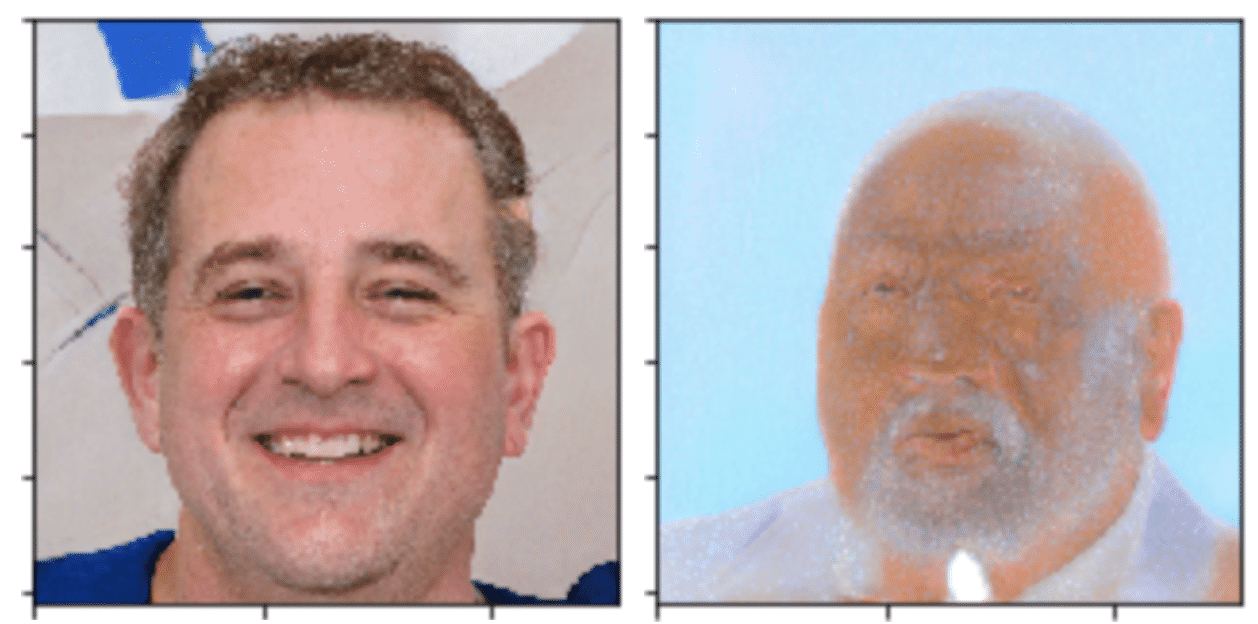}
\end{center}
   \caption{Mode collapse when the starting point is perturbatively away from zero}
\label{fig:mode-perturb}
\end{figure}

However, the understanding in terms of non-normalizable norm would indicate that by perturbing the zero vector slightly such that it can be normalized, the mode instability should be greatly ameliorated. As shown in Figure \ref{fig:mode-perturb}, even when the starting point is a normalizable latent vector, it still leads to mode collapse.

\begin{figure}[h]
\centering
\begin{subfigure}{.25\columnwidth}
  \centering
  \includegraphics[width=.9\linewidth]{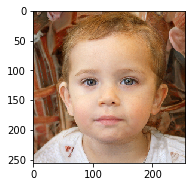}
  \caption{}
  \label{fig:mode-displace0}
\end{subfigure}%
\begin{subfigure}{.25\columnwidth}
  \centering
  \includegraphics[width=.9\linewidth]{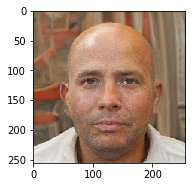}
  \caption{}
  \label{fig:mode-displace1}
\end{subfigure}%
\begin{subfigure}{.25\columnwidth}
  \centering
  \includegraphics[width=.9\linewidth]{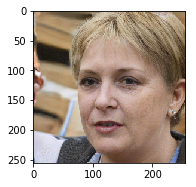}
  \caption{}
  \label{fig:mode-var0}
\end{subfigure}%
\begin{subfigure}{.25\columnwidth}
  \centering
  \includegraphics[width=.9\linewidth]{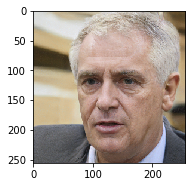}
  \caption{}
  \label{fig:mode-var1}
\end{subfigure}%
\caption{Sufficient perturbation resolves mode collapse. a: Starting from nonzero uniform. b: Starting from nonzero uniform, more male. c: Starting with larger variance. d: Starting with larger variance, more male}
\label{fig:mode-displace-var}
\end{figure}

When the starting point has a large enough variance or is displaced away from zero, the situation is indeed ameliorated. As shown in Figure \ref{fig:mode-displace-var}.

Figure \ref{fig:mode-displace0} corresponds to the starting point being a nonzero vector with uniform values and Figure \ref{fig:mode-displace1} is the more male version, free from mode collapse. Figure \ref{fig:mode-var0} corresponds to the starting point being a nonzero vector with normal distribution around zero with variance comparable to the variance of the vector describing the traversing direction. The more male version in Figure \ref{fig:mode-var1} is also free from mode collapse.

\begin{figure}[h]
\centering
\begin{subfigure}{.5\columnwidth}
  \centering
  \includegraphics[width=.9\linewidth]{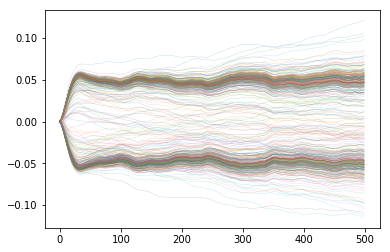}
  \caption{Starting from zero vector}
  \label{fig:mode-latent1}
\end{subfigure}%
\begin{subfigure}{.5\columnwidth}
  \centering
  \includegraphics[width=.9\linewidth]{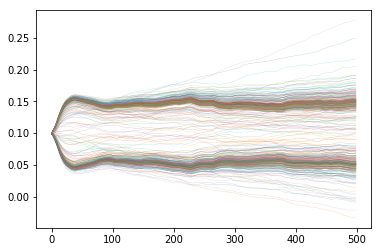}
  \caption{Starting from nonzero uniform vector}
  \label{fig:mode-latent2}
\end{subfigure}
\begin{subfigure}{.5\columnwidth}
  \centering
  \includegraphics[width=.9\linewidth]{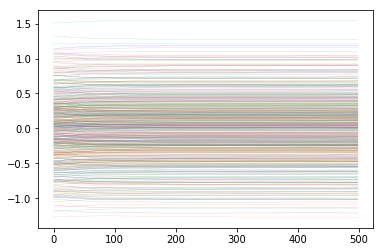}
  \caption{Starting from normal distribution}
  \label{fig:mode-latent3}
\end{subfigure}%
\begin{subfigure}{.5\columnwidth}
  \centering
  \includegraphics[width=.9\linewidth]{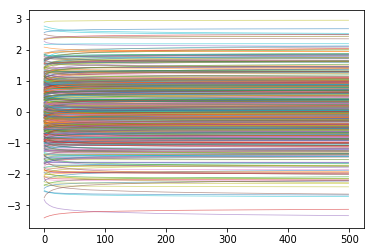}
  \caption{Starting from normal image}
  \label{fig:mode-latent4}
\end{subfigure}%
\caption{Latent vector change}
\label{fig:mode-latent-evolv}
\end{figure}

At this point, a more abstract characterization of the mode collapse phenomenon can shed more light. We first visualize how each of the 512 entries of the latent vector change during the latent space traversing, as shown in Figure \ref{fig:mode-latent-evolv}.

\begin{figure}[h]
\centering
\begin{subfigure}{.5\columnwidth}
  \centering
  \includegraphics[width=.9\linewidth]{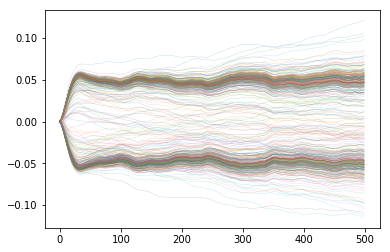}
  \caption{Starting from zero vector}
  \label{fig:mode-act1}
\end{subfigure}%
\begin{subfigure}{.5\columnwidth}
  \centering
  \includegraphics[width=.9\linewidth]{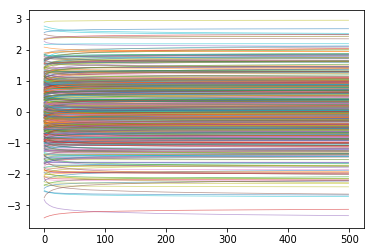}
  \caption{Starting from normal image}
  \label{fig:mode-act4}
\end{subfigure}%
\caption{Neuron activation change}
\label{fig:mode-act-evolv}
\end{figure}

Figure \ref{fig:mode-latent1} is the problematic case where traversing starts from zero latent vector, whereas Figure \ref{fig:mode-latent2}, \ref{fig:mode-latent3}, \ref{fig:mode-latent4} correspond to cases where the starting latent vector is nonzero and will not suffer from mode collapse. Notice the stark contrast that for the problematic case, the latent vector values concentrate in two extremes whereas for the cases free from mode collapse, the latent vector entries are distributed in a relatively more uniform way. This qualitative difference in terms of latent vector distribution underlies the difference in robustness against mode collapse. Figure \ref{fig:mode-act-evolv} shows the dynamics of neuron activations in the auxiliary network during training, which displays the same pattern as for the dynamics of latent vector, where the problematic case corresponds to a distribution that concentrates on extreme values instead of uniformly. Namely, for the problematic case with mode collapse, the underlying reason is that both the latent vector and the neuron activations are saturated, whereas for normal cases they are not.

\begin{figure}[h]
\centering
\begin{subfigure}{.5\columnwidth}
  \centering
  \includegraphics[width=.9\linewidth]{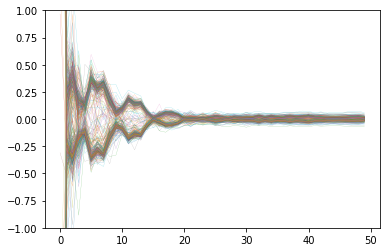}
  \caption{Starting from nonzero uniform vector}
  \label{fig:mode-ft-uni}
\end{subfigure}%
\begin{subfigure}{.5\columnwidth}
  \centering
  \includegraphics[width=.9\linewidth]{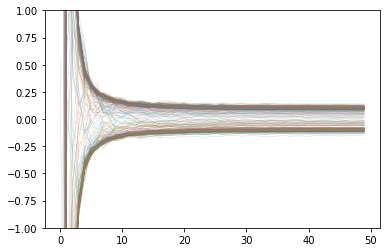}
  \caption{Starting from normal image}
  \label{fig:mode-ft-normal}
\end{subfigure}
\begin{subfigure}{.5\columnwidth}
  \centering
  \includegraphics[width=.9\linewidth]{fig/ft-var.png}
  \caption{Starting with larger variance}
  \label{fig:mode-ft-var}
\end{subfigure}%
\begin{subfigure}{.5\columnwidth}
  \centering
  \includegraphics[width=.9\linewidth]{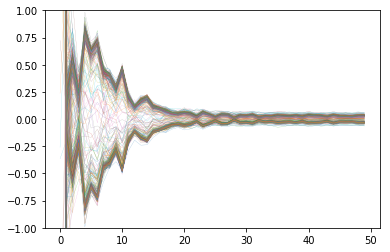}
  \caption{Starting with zero}
  \label{fig:mode-ft-zero}
\end{subfigure}%
\caption{Fourier transform of neuron activation change}
\label{fig:mode-ft}
\end{figure}

If one observes closely, apart from saturation, the problematic case with mode collapse also gives rise to small amount of oscillations in Figure \ref{fig:mode-latent-evolv} and \ref{fig:mode-act-evolv} in terms of the dynamics of latent vector and neuron activations. To characterize the oscillation, the Fourier transform of the evolution of neuron activation is provided in Figure \ref{fig:mode-ft}.

Figure \ref{fig:mode-ft-zero} corresponds to the problematic case of traversing from zero with most severe mode collapse, whereas Figure \ref{fig:mode-ft-uni}, \ref{fig:mode-ft-normal}, \ref{fig:mode-ft-var} are cases with ameliorated mode collapse and oscillation. The oscillation is reflected in the Fourier spectrum in the sense that larger oscillation corresponds to larger amplitude of high frequency components. Figure \ref{fig:mode-ft} is consistent with Figure \ref{fig:mode-act-evolv} and traces back to the ease of saturation when the starting point of traversing is near zero.

To solve the mode collapse, the simplest solution is to traverse $\mathcal{W}$ space instead of $\mathcal{Z}$ space, since W space is more separable and the direction obtained through the auxiliary network is less noisy. This is confirmed in Section \ref{preliminary} where traversing in $\mathcal{W}$ space near zero is free from mode collapse problem.

\section{Conclusion and outlook}

We implement an image editing tool based on Style-GAN with good qualitative and quantitative performance. Our method can be easily extended to content-aware image editor based on other GANs (e.g. full-body pose, accessories, background etc.). Moreover, the nonlinear network can be further optimized to improve feature disentanglement and to exploit the flexibility for tuning features that are otherwise too subtle to define. Our implementation also provides a concrete setting for the mode collapse problem and our characterization of the problem may provide insight on mode collapse problems in more general settings.

\newpage
{\small
\bibliographystyle{ieee}
\bibliography{egbib}
}

\end{document}